\documentclass[letterpaper, 10 pt, conference]{ieeeconf}
\IEEEoverridecommandlockouts
\usepackage[sorting=none]{biblatex}
\bibliography{references}
\usepackage{color, colortbl}

\usepackage{amsmath,amssymb,amsfonts}
\usepackage{graphicx}
\usepackage{textcomp}
\usepackage{xcolor}
\usepackage{hyperref}
\usepackage{algorithm}
\usepackage{algpseudocode}
\usepackage{tabularx}
\usepackage{multirow}
\usepackage{soul}

\def\BibTeX{{\rm B\kern-.05em{\sc i\kern-.025em b}\kern-.08em
    T\kern-.1667em\lower.7ex\hbox{E}\kern-.125emX}}

\makeatletter
\newcommand*\titleheader[1]{\gdef\@titleheader{#1}}
\AtBeginDocument{%
  \let\st@red@title\@title
  \def\@title{%
    \bgroup\normalfont\large\centering\@titleheader\par\egroup
    \vskip1.5em\st@red@title}
}
\makeatother

\title{
Correcting Motion Distortion for LIDAR HD-Map Localization \\
}

\titleheader{\scriptsize This work has been submitted to the IEEE for possible publication. Copyright may be transferred without notice, after which this version may no longer be accessible.}
\author{Matthew McDermott$^{1}$ and Jason Rife$^{2}$% <-this % stops a space
\thanks{$^{1}$Matthew McDermott is a student in the Mechanical Engineering Ph.D. program at Tufts University in Medford, MA. He works in the Automated Systems and Robotics Laboratory (ASAR) with Dr. Jason Rife. He received his B.S. and M.S. degrees in Mechanical Engineering at Tufts University,
        {\tt\small matthew.mcdermott@tufts.edu}}%
\thanks{$^{2}$Jason Rife is a Professor and Chair of the Department of Mechanical Engineering at Tufts University in Medford, Massachusetts. He directs the Automated Systems and Robotics Laboratory (ASAR), which applies theory and experiment to characterize integrity of autonomous vehicle systems. He received his B.S. in Mechanical and Aerospace Engineering from Cornell University and his M.S. and Ph.D. degrees in Mechanical Engineering from Stanford University.
        {\tt\small jason.rife@tufts.edu}}%
}

\begin{document}
\maketitle

\begin{figure*}[h]
    \includegraphics[width=\textwidth]{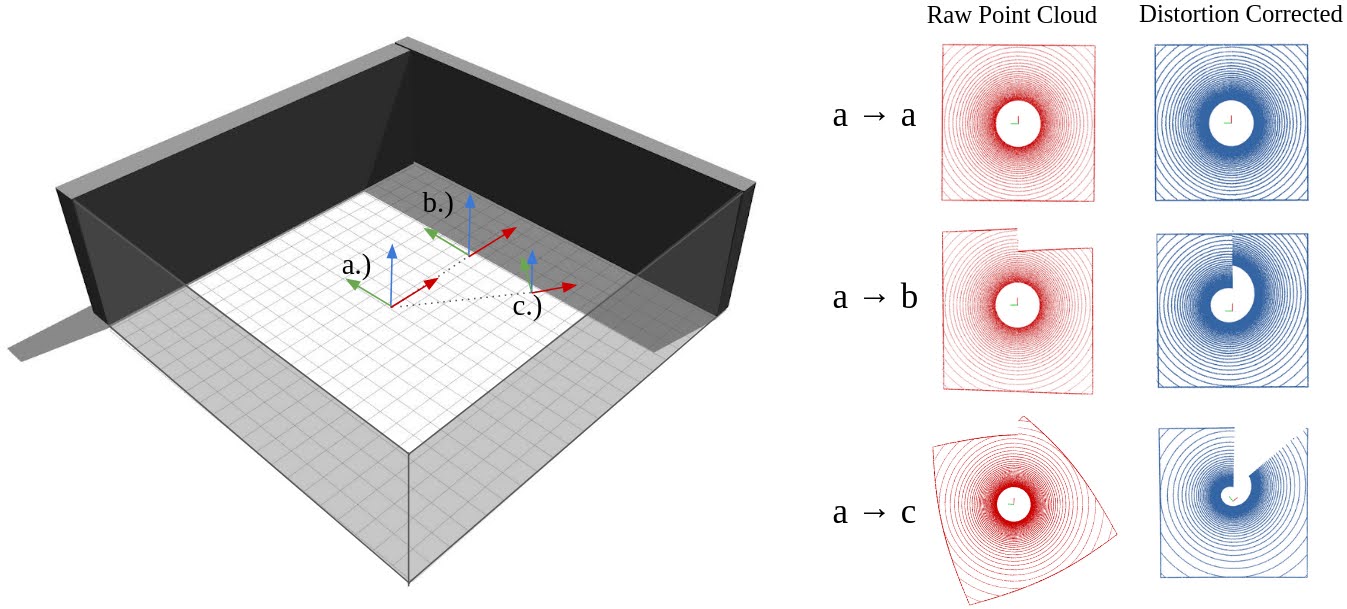}
  \caption{Simple test scene to study motion-distortion correction. The isometric room view shows three possible LIDAR locations, labeled a, b, and c.  At each location, the orientation of the LIDAR unit is described by a set of orthonormal basis vectors. For each configuration, assume the LIDAR beam begins aligned with the red arrow and rotates counterclockwise about the vertical (blue) axis. During the scan, the LIDAR unit either remains stationary (a~→~a), undergoes forward linear motion (a~→~b), or undergoes composite translation and rotation (a~→~c).  In each case, the LIDAR beam spins $360^\circ$ in the frame of the LIDAR stator, which itself moves, resulting in distorted raw point clouds (viewed from above, shown in red).  Compensating for stator motion, the raw image can be transformed into room-fixed coordinates (shown in blue), where the square shape of the room is recovered.}
  \label{fig:threeDistortions}
\end{figure*}

\begin{abstract}
% In this paper we show how jointly solving for relative motion distortion between LIDAR scans can be used to provide a bound for position error. This is because, without an \textit{a priori} model of the environment, there are many potential combinations of ground truth accelerations and world geometries that could have produce the point clouds observed by the sensor.
% In our proposed method, we extend NDT to solve for the relative difference in the velocity of the platform between scans, in addition to the rigid transform required to align their origins. We provide experiments to demonstrate how our method of distortion compensation achieves lower Chamfer Cistance (CD) than point clouds rectified by both pinning distortion compensation parameters directly to position states (as in VICP) and also pre-correcting distortion via a model of platform motion (Kalman-NDT).     
% We conduct experiments on real and simulated data to demonstrate the utility of these bounds.      

Because scanning-LIDAR sensors require finite time to create a point cloud, sensor motion during a scan warps the resulting image, a phenomenon known as \textit{motion distortion} or \textit{rolling shutter}.  %Motion distortion causes systematic error for standard LIDAR scan-matching techniques. 
Motion-distortion correction methods exist, but they rely on external measurements or Bayesian filtering over multiple LIDAR scans.
%A scanning LIDAR sensor will misrepresent its surrounding environment if it is in motion while recording data. This phenomenon, known as \textit{motion distortion}, will warp the geometry of measured objects thereby causing traditional scan matching techniques to converge on incorrect solutions. 
In this paper we propose a novel algorithm that performs snapshot processing to obtain a motion-distortion correction.  Snapshot processing, which registers a current LIDAR scan to a reference image without using external sensors or Bayesian filtering, is particularly relevant for localization to a high-definition (HD) map.  Our approach, which we call Velocity-corrected Iterative Compact Ellipsoidal Transformation (VICET), extends the well-known Normal Distributions Transform (NDT) algorithm to solve jointly for both a 6~Degree-of-Freedom (DOF) rigid transform between two LIDAR scans and a set of 6DOF motion states that describe distortion within the current LIDAR scan. Using experiments, we show that VICET achieves significantly higher accuracy than NDT or Iterative Closest Point (ICP) algorithms when localizing a distorted raw LIDAR scan against an undistorted HD Map. 
% We also demonstrate how VICET estimates motion states that can be used to bound scan-to-scan registration errors when initializing with a distorted keyframe.
We recommend the reader explore our open-source code and visualizations at \url{https://github.com/mcdermatt/VICET}, which supplements this manuscript.
\end{abstract}

% \begin{IEEEkeywords}
% component, formatting, style, styling, insert
% \end{IEEEkeywords}

\section{Introduction}
A scanning LIDAR records data by  sweeping an array of laser-ranging sensors across a scene, typically completing a full sweep once every 50-200 ms. 
% typically at a rate of 5-15 Hz. 
%Objects in the same sweep are recorded at different times depending on their positions. 
Even modest motion of the sensor unit during the sweep creates discernible distortion in the scanned image. By contrast, instantaneous sweeps are assumed in conventional scan-matching algorithms such as Iterative Closest Point (ICP) \parencite{ICP}, the Normal Distributions Transform (NDT) \parencite{biber2003normal}, and their variants \parencite{GICPCostFactors, 3DICET, stereoNDT, D2DNDT}. Although the instantaneous-sweep, distortion-free approximation is reasonable for a static platform observing a static scene, distortion is usually present in LIDAR data captured by a moving platform.
%in most for frames that are bent or stretched challenge the scan-registration process and introduce systematic biases into the solution. %however, range measurements no longer share a common origin when the sensor is in motion.
%Instead,  motion of the platform will cause points to move with the frame of sensor during the period of the scan. 
%however, even modest translational or rotational motion appear as distortions of the scanned image.Failing to account for this change in frame will cause surfaces in the recorded point cloud to appear bent and stretched relative to their true shapes. 
Such motion distortion effects impact all LIDAR registration algorithms, whether keypoint-based \parencite{LOAM, deepICPcovEstimation}, distributions-based \parencite{3DICET, D2DNDT}, or ML-based \parencite{lonet, unsupervisedDNN}. 

Because motion distortion is a ubiquitous issue for LIDAR scan matching, we compare
%This is because it embeds error into the clouds themselves that can not be quantified without additional information about the true geometry of the scene. Before describing our proposed algorithm, it is useful to first introduce 
existing techniques for combating motion distortion and discuss their strengths and limitations. One simple method of alleviating motion-distortion bias is to augment raw LIDAR data with external sensor measurements that describe platform velocity and/or angular velocity, for example, from Global Navigation Satellite System (GNSS), Interial Navigation System (INS), or wheel odometry data.  Repositories of LIDAR benchmark data often publish pre-rectified point clouds in which distortion effects have already been mitigated using external sensors \parencite{ford, pixset, nuscenes}. 
Although distortion can be mitigated somewhat, external sensors also introduce new distortion due to sensor noise. This noise may be substantial. For example, GNSS velocity estimates in clear-sky conditions are roughly 3-5 cm/s (one-sigma) \parencite{misra2004global}) and, in high-multipath environments like urban canyons, are much worse \parencite{multipath}. By comparison, velocity estimates generated from the LIDAR data, itself, can achieve higher accuracy, often better than 1~cm/s. 

A second method compensates for motion distortion by estimating platform motion with batch or sequential filtering, for instance, using a Kalman Filter \parencite{inui2017distortion}. A filter can suppress the noise from an external sensor, but the filter also introduces lag~\cite{franklinPowell}, resulting in suppression of high-frequency velocity-variations and making velocity corrections difficult to match to the correct LIDAR time step, particularly when the platform changes speed rapidly. 

A third approach enhances the performance of estimation filters through tighter coupling with the LIDAR image-generation process. For instance, Setterfield exploits factor-graph optimization to break each LIDAR image into highly localized SURF features, each of which represents a sample interval short enough ($<$1~ms) to be essentially distortion free \cite{setterfield}. This approach, however, extracts a small number of time-aligned features and discards the remaining LIDAR points; moreover, the approach relies on highly textured terrain and struggles when viewing large smooth surfaces  \cite{shimojo1989occlusion}. Given these limitations, other tightly-coupled estimators like VICP \parencite{vicp} and LOAM \parencite{LOAM} mitigate motion distortion for the entire point cloud by estimating a time-varying transform with linear scaling across the interval of a full scan. Both VICP and LOAM obtain velocity estimates from the LIDAR data directly, by analyzing data in batches and constructing a local submap in the process. These tightly-coupled algorithms do not function well unless processing a long series of sequential frames.  The issue is magnified for VICP and LOAM, since motion distortion in the initial images of the sequence can significantly warp the local submap and thereby corrupt early velocity estimates.
%Though useful in the odometry application, both require access to a local submap onto which new LIDAR data can be aligned. In other words, these approaches are only capable of registering a new distorted scan against an existing keyframe if the keyframe is pre-aligned with body frame of the sensor. 
%This limitation prevents LOAM's motion correction from being used for scan to HD Map localization where both position and velocity of the sensor are unknown. LOAM implementations generally start a new sub-map from the first available point cloud scan each time the algorithm is initialized. Error is introduced if this first scan comes from a moving platform, because subsequent LIDAR scans will be "unwrapped" to match the distorted reference, rather than the true geometry of the scene, potentilly over or under estimating platform motion as a result. 
%This is made worse by the fact that LOAM does not provide any metric of it's uncertainty regarding a registration estimate.  

We seek a new approach for LIDAR-based distortion correction that does not rely on external measurements or an estimator, such that snapshot processing is possible, performing registration using only a reference image and the current LIDAR scan. This is particularly important for map-based navigation, where instantaneous position is estimated by matching a single scan to a pre-existing map.  

We also seek a distortion-correction methodology that produces a meaningful prediction of measurement uncertainty (e.g. a covariance matrix that quantifies the error for aligning two point clouds).  Even sequential-estimator based distortion correction methods, like VICP and LOAM, do not provide meaningful uncertainty quantification, in part because they use scan-matching methods for which accuracy is difficult to predict. Development of robust error bounds is an important aspect of safety-critical navigation systems such as autonomous passenger vehicles \parencite{kigotho2022lane}. 

We seek to achieve both goals by reformulating the Normal Distribution Transform (NDT)~\cite{biber2003normal}, a scan-matching algorithm designed for registration of scan pairs and readily adapted for scan-to-map matching. 
% Numerous works explore registration uncertainty in NDT \parencite{D2DNDT, 2DICET, 3DICET, taylorICPcov}.
Uncertainty quantification methods exist for NDT, dating back to Stoyanov \parencite{D2DNDT}. Although this early work has not been widely adopted due to its difficult implementation, we recently reformulated NDT (via an algorithm called ICET \parencite{3DICET}) to  streamline error estimation while also excluding a significant source of error related to the distinction between random noise and structural patterns of voxel distributions \parencite{3DICET,2DICET}. 
Our primary contribution in this paper is to modify NDT by adding motion-distortion compensation, in order to enhance registration accuracy. 
% \hl{while bounding systematic-bias errors due to unknown keyframe distortion.}
% In parallel, others have worked to model error distributions for other types of point cloud registration algorithms including ICP \parencite{taylorICPcov} and landmark-based methods \parencite{joerger2022uncertainty}. As LIDAR sensor quality improves to produce point clouds of higher angular resolution and less range noise, errors from motion distortion become more dominant in the localization process. Therefore, it is imperative to develop techniques for localization that are both accurate in the presence of motion distortion and transparent about their uncertainty. 
%In this paper we demonstrate how our proposed algorithm is capable of outperforming exiting scan registration methods in scan-to-HD Map localization, and show how our technique is capable of bounding its own solution error in cases where the distortion in a keyframe is unknown.  

% Such an approach may allow LIDAR to be used in highly dynamic applications such as high speed drone flight.   

The remainder of the paper is organized as follows. In section II we refine our definition of motion distortion and demonstrate how relative distortion between two scans can produce ambiguity in the scan-registration process. In section III,  we formulate our new algorithm, VICET, and discuss specific implementation details. In section IV we compare VICET performance against benchmark algorithms of NDT and ICP. In section V we discuss the results of our experiment and explore avenues of future work. Finally, in section VI we summarize our key findings. 

% \newpage

\section{Motion Distortion Example}

% \begin{figure}[h]
% \centering
% \includegraphics[width=3.2in]{graphics/threeposesboxv2.png}
% % \includegraphics[width=3.5in]{graphics/threedistortionsv2.png}
% \includegraphics[width=3.5in]{graphics/threedistortionsv3.jpg}
% \caption{Simple test scene with three pairs of point clouds before and after motion distortion correction. 
% % \textcolor{red}{TODO: recolor this so top row is red, middle row is blue, bottom row is green so it can more easily be related to figure 2}
% }
% \label{fig:threeDistortions}
% \end{figure}

Before describing our solution, it is helpful to develop a better understanding of motion distortion through a simple simulation.  Consider a mechanically spinning LIDAR unit placed inside a rectangular room, as illustrated in Fig. \ref{fig:threeDistortions}.  A LIDAR image (or \textit{scan}) is generated each time the LIDAR spins $360^\circ$ about the LIDAR's vertical ($z$) axis.  Define the LIDAR scan to begin when the rotating beam aligns with the positive $x$ direction in a coordiante system attached to the LIDAR stator, which we will label the \textit{body} frame.  In the figure, starting and ending locations of the LIDAR are indicated by vector-bases where the spin axis is shown in blue and the scan-initiation axis is shown in red.  Now consider three cases, one in which the sensor remains stationary during the scan (starting and ending at the pose labeled \textit{a}), a second in which the sensor undergoes pure translation (starting at \textit{a} and ending at \textit{b}), and a third in which the sensor undergoes both translation and rotation (starting at \textit{a} and ending at \textit{c}).  Each of the three cases creates a distinct LIDAR point cloud shown, from an aerial view, in a column labeled ``Raw Point Cloud'' (red).
%The resulting raw point cloud (drawn in red) portrays the scene as it appears from the perspecive of the sensor-- with each shooting direction of the beam potentially occuring with the body of the sensor at different point in space as the device moves throughout the scene. In in the top example (denoted with $a\,\to\,a$), the LIDAR unit is placed at position $a$ and remains static for the duration of the scan. For the static case, there is no motion distortion, so the corrected cloud (drawn in blue) for this case looks identical. In the second case  (denoted $a\,\to\,b$), the LIDAR unit begins in position $a$ and moves forward at constant velocity for the duration of the scan ending at position $b$. Here the raw (red) point cloud contains clear evidence of motion distortion in the forward direction. In the final case (denoted $a\,\to\,c$), the sensor exhibits more complex motion with translation in two axis as well as rotation about it's z axis in the opposite direction as the scanning beam rotates. 
In all cases, a circle appears in the middle of the point cloud, reflecting an elevation cutoff, with the simulated LIDAR unit unable to generate samples below $30^\circ$ under the horizon.  The point clouds were generated assuming a uniform rate of motion between the start and end points.  Importantly, the raw images are distorted (except in the stationary case), such that the walls of the room do not form a perfect square.  Accounting for platform motion, however, points can be shifted from body-frame coordinates to world-frame coordinates. Unwarped point clouds are shown in Fig. \ref{fig:threeDistortions} as a column of images (blue) labeled ``Distortion Corrected.''  Importantly, these distortion-corrected images all recover the correct room shape, bounded by a square wall.  In the last case ($a \rightarrow c$), a wedge of missing data appears because the LIDAR stator rotates clockwise, opposite the counterclockwise rotation of the LIDAR rotor, such that the entire room is not visualized during a single scan.

%Motion effects can be visualized by comparing the ``Distortion Corrected'' scan to the ``Raw Point Cloud.''  In the first case, when the simulated LIDAR unit is stationary ($a \rightarrow a$), the corrected scan and raw scan are identical.  By contrast, in the pure translation case ($a \rightarrow b$), then a gap appears along one of the room's walls, reflecting that the sensor moves toward the wall during the duration of the scan (with the LIDAR rotating in the counterclockwise direction). In the combined translation and rotation case ($a \rightarrow c$), then the rotation of the sensor platform ($45^{\circ}$ clockwise, from above) acts opposite to the scanning rotation, such that a full sweep of the laser beam (in platform coordinates) covers only $315\circ$ of the room.  The result is a missing wedge of data in the frame of the world, as seen from the distortion corrected image ($a \rightarrow c$ case).

For map-matching applications, pose is estimated relative to a reference image. For instance, the reference image might be constructed by registering a series of sequential scans from a moving LIDAR, in order to create a mosaic image or high-definition (HD) map. The performance of this registration operation is greatly enhanced if the current scan (captured in the frame of the LIDAR) can be unwarped (converted to the world frame) before registration to the map.  In the case of the rectangular room of Fig. (\ref{fig:threeDistortions}), the map will look very much like the scan for the static case ($a \rightarrow a$). If a new LIDAR image were generated during translation ($a \rightarrow b$) or combined translation and rotation ($a \rightarrow c$) the resulting raw image (red) will be harder to align with a map of the room than the corresponding distortion-corrected image (blue).

The main focus of this paper is that distortion correction is possible using only the current LIDAR scan and a map.  If LIDAR motion is not known, then we can infer the motion by registering the current image to a map while also testing various unwarping transformations, in order to determine the combination of registration and unwarping that results in best alignment. We develop this idea for simultaneous estimation of pose and motion-correction states in subsequent sections. 

%Conversely, the platform velocity and angular velocity can be \textit{inferred} by reversing the process and determining what unwarping operation converts the raw data to look most like the map.  This is the basis of our algorithm described in the next section, which attempts to solve for the velocities that best unwarp a raw point cloud while simultaneously solving for the unknown pose of the LIDAR unit relative to a map.

We intentionally restrict our analysis to cases where the reference image is an undistorted map.  In concept, the process of aligning and unwarping images could also be performed on sequential images, for instance to implement LIDAR odometry.  If the reference image were distorted, our estimated motion-distortion correction would reflect relative velocity and angular velocity (changes between scans) but not absolute velocity and angular velocity (motion relative to the ground).  In other words, our approach does not correct distortion in the reference image.

\section{Implementation}

This section develops a snapshot minimum-squared error (MSE) estimator, which infers both the relative-pose (translation and rotation) and motion-distortion parameters for a current LIDAR scan by comparison to an undistorted reference map.  We call the algorithm Velocity-corrected Iterative Compact
Ellipsoidal Transformation (VICET).

\subsection{Algorithm Formulation}

\begin{figure}[h]
\centering
\includegraphics[width=2.0in]{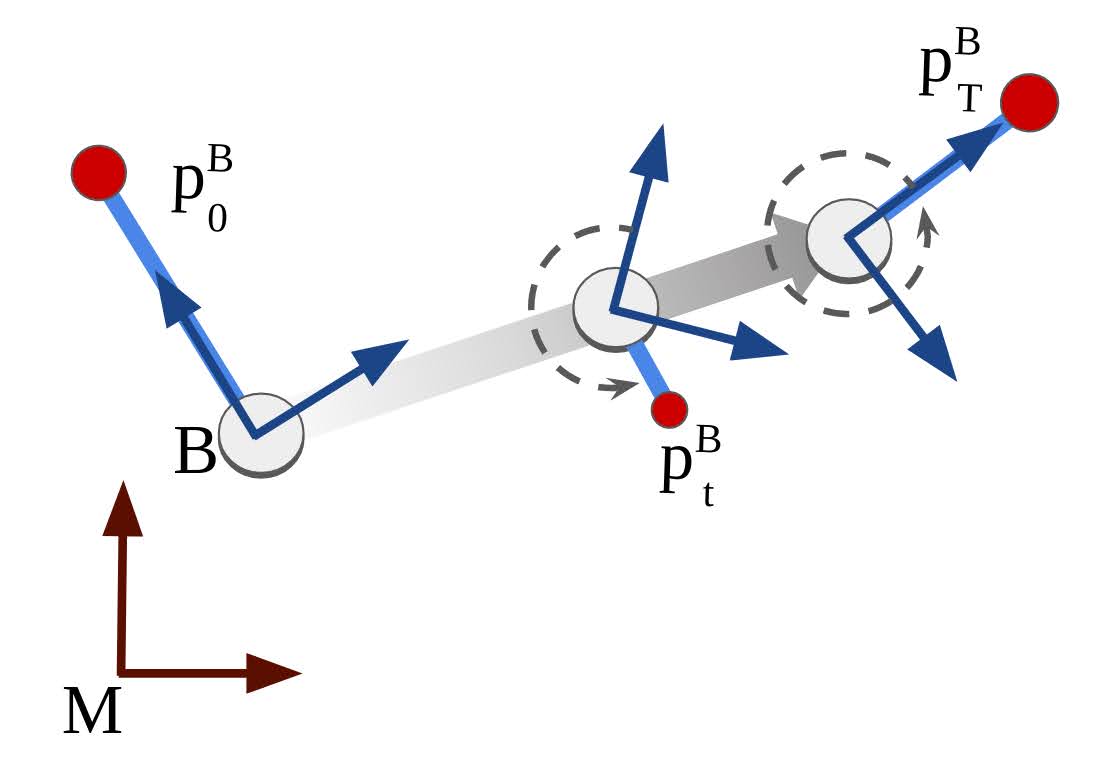}
\caption{The frame of the LIDAR stator (or \textit{body}) B moves relative to the map frame M.  For illustration purposes, the bases are shown in 2D, viewed from above. 
 The progress of time is indicated by the shaded gray arrow, with the LIDAR beam recording measurements (red dots) as it sweeps a full circle between time $t=0$ and time $t=T$.}
\label{fig:fromulationFrames}
\end{figure}

To start, we define two relevant reference frames: one associated with the map and a second associated with the LIDAR stator. We characterize the map frame by an origin $Q_M$ and a coordinate system defined by the basis $M$.  We define the LIDAR stator frame, which we also called the \textit{body} frame, with a reference point $Q_B$ and a body-fixed coordinate system $B$.  The body frame translates and rotates relative to the map frame over time, as shown in Fig.~\ref{fig:fromulationFrames}.  Although the LIDAR rotor spins about a fixed axis of B, we do not define a distinct reference frame for the rotor, since the LIDAR system converts measurements to the stator frame~B using encoder data.

A map matching algorithm seeks to find the rigid transform that aligns the raw LIDAR scan to the map at the beginning of the scan, at time $t=0$.  The motion states are inferred to unwarp the LIDAR scan given its motion between the beginning of the scan and the end of the scan, which occurs at $t=T$.  Using a zero-order hold model, we approximate that the velocity and angular velocity of $B$ relative to $M$ remains constant for times $t\in[0,T]$.  Now consider a LIDAR measurement vector $\mathbf{p}^B_t$.  The vector $\mathbf{p}^B_t$ describes the position of a feature in the world (or on the map) relative to the body-frame reference point $Q_B$ using body-fixed coordinates (specified by the trailing superscript $B$) at a time $t$ (specified by the trailing subscript).  The following transformation converts this measurement to $\mathbf{p}^M_t$, which expresses the same measurement in map coordinates relative to the map origin $Q_M$:

\begin{equation}\label{eq:B02Map}
    \mathbf{p}^{M}_t = {}^M\mathbf{R}^{B}_t \mathbf{p}^{B}_t + \mathbf{m}^M_t
\end{equation}

\noindent Here the rotation matrix ${}^M\mathbf{R}^{B}_t$ converts from the $B$ basis to the $M$ basis at time $t$.  The translation $\mathbf{m}^M_t$ expresses the vector from $Q_M$ to $Q_B$ at time $t$, in map-fixed coordinates.  

As a next step, we decompose the terms in (\ref{eq:B02Map}) to relate them to initial states and change states.  This decomposition is easiest to develop for the translation term $\mathbf{m}^M_t$, which is linear.  Given the zero-order hold assumption, we can rewrite the translation using the lever rule to interpolate between an initial translation $\mathbf{m}^M_0$ and a final translation $\mathbf{m}^M_T$.  The lever rule can be written

\begin{equation}\label{eq:lever}
    \mathbf{m}^M_t =\mathbf{x}_0 + s\Delta\mathbf{x}
\end{equation}

\noindent This formulation introduces the following shorthand notation: $\mathbf{x}_0=\mathbf{m}^M_0$, $\Delta\mathbf{x}=\mathbf{m}^M_T-\mathbf{m}^M_0$, and $s=t/T$. The scaled time $s$, derived from the LIDAR time tag for each measurement, is normalized such that $s\in[0,1]$.  The variables $\mathbf{x}_0$ and $\Delta\mathbf{x}$ are identified specifically because these are states that the estimator will infer.

Just as initial and change states are defined for translation, similar states must be defined for rotation.  In our implementation, we use an array of three Euler angles $\mathbf{\Theta_0}$ to describe the initial orientation of the body frame.  The rotation matrix is constructed as a function $f$ of the Euler angles $\mathbf{\Theta_0}$.

\begin{equation}\label{eq:RotMatrix}
    {}^M\mathbf{R}^{B}_0 = f(\mathbf{\Theta_0})
\end{equation}

\noindent The orientation change is described by a set of three angles $\Delta\mathbf{\Theta}$.  The final rotation matrix at time $T$ can be related, as follows, to the initial rotation matrix and $\Delta\mathbf{\Theta}$.

\begin{equation}\label{eq:RotChange}
    {}^M\mathbf{R}^{B}_T({}^M\mathbf{R}^{B}_0)^{-1} = f(\mathbf{\Delta\Theta})
\end{equation}

Compiling the initial and change states together, we can define a combined state vector $\mathcal{X}\in\mathbb{R}^{12\times1}$, which we will estimate by comparing the LIDAR scan to a map. 

\begin{equation}\label{eq:newDefChi}
    \mathcal{X} = \begin{bmatrix}
        \mathbf{x}_0 \\  \Delta \mathbf{x} \\
        \mathbf{\Theta_0} \\
        \Delta\mathbf{\Theta}
    \end{bmatrix}
\end{equation}

To relate (\ref{eq:B02Map}) to the states in (\ref{eq:newDefChi}), we use a Taylor series expansion, in which we expand (\ref{eq:B02Map}) about a reasonable guess.  In implementation, we seed the initial position and Euler angles with a rough guess, and we seed the change parameters with zero values. The affine terms of the Taylor series are:

\begin{equation}\label{eq:Taylor}
    \mathbf{p}^{M}_t = ({}^M\mathbf{\hat{R}}^{B}_t +
    \delta{}^M\mathbf{R}^{B}_t)
    \mathbf{p}^{B}_t + \mathbf{\hat{m}}^M_t +  
    \delta \mathbf{m}^M_t.
\end{equation}

\noindent  Here we employ a hat notation to identify terms computed using the initial guess. Also, we introduce a $\delta$ notation to identify linear perturbations.

Because the translation term is linear, the translation perturbation can be obtained from the derivative of (\ref{eq:lever}):

\begin{equation}\label{eq:lever}
    \delta\mathbf{m}^M_t =\delta\mathbf{x}_0 + s\delta\Delta\mathbf{x}.
\end{equation}

The rotation-perturbation $\delta\mathbf{R}$ is obtained from the rotation-matrix derivative, which in the small angle limit is
%The result is a cross-product matrix multiplied by the original rotation matrix.  The cross-product matrix can be constructed (in the small angle limit) from a set of Euler angle corrections as follows.

\begin{equation}\label{eq:gMap}
    \delta{}^M\mathbf{R}^{B}_t = 
     [\delta\mathbf{\Theta}{\times}] \;\,
    {}^M\hat{\mathbf{R}}^{B}_t.
\end{equation}

\noindent Here $\delta\mathbf{\Theta}$ is an angular correction converted to a cross-product by the $[\_\_\times]$ operator.  This matrix, for a generic vector $\textbf{v}$ with elements $\{v1,v2,v3\}$ is:

\begin{equation}\label{eq:gMapB}
    [\mathbf{v}\times] =  \begin{bmatrix}
        0 & -v_3 & v_2 \\
        v_3 & 0 & -v_1 \\
        -v_2 & v_1 & 0
    \end{bmatrix}.
\end{equation}

Though Euler angles do not sum in general, summation is a good approximation after linearization, and so $\delta\mathbf{\Theta}$ can be written in terms of initial and change states:

\begin{equation}\label{eq:leverRot}
    \delta\mathbf{\Theta} =\delta\mathbf{\Theta}_0 + s\delta\Delta\mathbf{\Theta}
\end{equation}

\noindent Substituting (\ref{eq:lever})-(\ref{eq:leverRot}) into (\ref{eq:Taylor}), we obtain a linear system of equations relating the correction states to the LIDAR data:

\begin{equation}\label{eq:TaylorH}
    \mathbf{p}^{M}_t = {}^M\mathbf{\hat{R}}^{B}_t
    \mathbf{p}^{B}_t + \mathbf{\hat{m}}^M_t +
    \mathbf{H}_\mathcal{X}
    \delta \mathcal{X}. 
\end{equation}

\noindent Here the 
the state-perturbation vector  $\delta\mathcal{X}\in\mathbb{R}^{12\times1}$ is

\begin{equation}\label{eq:defDeltaChi}
    \delta\mathcal{X} = \begin{bmatrix}
        \delta\mathbf{x}_0 \\  \delta\Delta \mathbf{x} \\
        \delta\mathbf{\Theta_0} \\
        \delta\Delta\mathbf{\Theta}
    \end{bmatrix},
\end{equation}

\noindent and the Jacobian $\mathbf{H}_\mathcal{X}\in\mathbb{R}^{3\times12}$ is

\begin{equation}\label{eq:defHchi}
    {\mathbf{H_{\mathcal{X}}}} = 
    \begin{bmatrix}
        \mathbf{I} & s\mathbf{I} & -\mathbf{P} & -s\mathbf{P}  \\ 
    \end{bmatrix},
\end{equation}

\noindent where the cross-product is embedded in $\mathbf{P}=[{}^M\hat{\mathbf{R}}^{B}_t \, \mathbf{p}^{B}_t \times]$.

In order to invert equation (\ref{eq:TaylorH}) and obtain the state corrections (\ref{eq:defDeltaChi}), two important details remain.  One issue is data association, which is to say that the body-frame point described by $\mathbf{p}^{B}_t$ must be matched to the corresponding point $\mathbf{p}^{M}_t$ in the map data.  A second issue is that (\ref{eq:TaylorH}) contains only 3 equations, so several such systems must be solved simultaneously to  enable an independent solution for the twelve states in (\ref{eq:defDeltaChi}).  Moreover, to avoid condition-number deficiencies, points must be compared from a wide range of $s$ values (meaning a wide range of angles through the LIDAR rotational sweep); otherwise if the range of $s$ is narrow, the structure of (\ref{eq:lever}) undermines the independence of $\delta\mathbf{x}_0$ and $\delta\Delta\mathbf{x}$, and similarly the structure of (\ref{eq:leverRot}) undermines the independence of $\delta\mathbf{\Theta}_0$ and $\delta\Delta\mathbf{\Theta}$.  In the rest of this section, we assume a broad swathe of features are imaged across the full range of $s$, such that a well-condition set of equations can be obtained.  Thus, our main focus is data association.

To address the data association problem, we use a voxel-based strategy, as employed by the popular NDT algorithm and our variant ICET.  
In these algorithms, point clouds are situated within a three-dimensional grid, consisting of volume elements called \textit{voxels}. 
If the voxel grid is defined in frame M, it is trivial to assign each LIDAR point from the HD-map to the voxel containing it.  Points from the current scan can also be associated with voxels, after the scan is transformed to frame M using the initial guess. An improved registration can then be inferred by aligning the distribution of map points to the distribution of current-scan points within each voxel. 
Ultimately, the grid enforces spatial associations, thereby avoiding association problems that arise in ICP, LOAM and other methods in the form of ambiguities caused by incorrect matches of points (or extracted features) between the map and the current scan.

To compare distributions of points in a given voxel, we take the expected value of (\ref{eq:TaylorH}) across all points in the voxel, which gives the following.

\begin{equation}\label{eq:TaylorMean}
    E\big(\mathbf{p}^{M}_t\big) = E\big({}^M\mathbf{\hat{R}}^{B}_t
    \mathbf{p}^{B}_t \big)+ E\big(\mathbf{\hat{m}}^M_t \big)+
    E\big(\mathbf{H}_\mathcal{X}
    \delta \mathcal{X}\big) 
\end{equation}

On the right side of the equation, the rotation matrix ${}^M\mathbf{\hat{R}}^{B}_t$, the translation vector $\mathbf{\hat{m}}^M_t$, and the state-correction vector $\mathcal{X}$ are the same for all points in the cloud (and therefore also for all points in voxel). Accordingly, these three terms can be moved outside the expected-value operator.  

\begin{equation}\label{eq:TaylorMu1}
    E\big(\mathbf{p}^{M}_t\big) = {}^M\mathbf{\hat{R}}^{B}_t E\big(\mathbf{p}^{B}_t \big)+ \mathbf{\hat{m}}^M_t+
    E\big(\mathbf{H}_\mathcal{X}
    \big) \delta \mathcal{X}
\end{equation}

\noindent In this equation, the matrix $\mathbf{H}_\mathcal{X}$  depends on $s$ and $P$, which have slightly different values for each point in the voxel.  Those variations are small in practice, and a very good approximation is obtained by defining ${}^{(j)}\mathbf{\tilde{H}}_\mathcal{X}$, which is equivalent to $\mathbf{H}_\mathcal{X}$ except that, in (\ref{eq:defHchi}), the $s$ value is determined for the midpoint of voxel $j$ and $\mathbf{P}$ is set to $\mathbf{P}=[{}^M\hat{\mathbf{R}}^{B}_{mid} \; {}^{(j)}\mathbf{\mu}^{B} \; \times]$, with the rotation matrix also evaluated at the voxel midpoint. Using this simplification, we can closely approximate (\ref{eq:TaylorMu1}) by setting ${}^{(j)}\mathbf{\tilde{H}}_\mathcal{X}\approx E\big(\mathbf{H}_\mathcal{X} \big)$.  The remaining two expected-value operations are averages across 
LIDAR points in a given voxel $j$. Accordingly, we invoke the $j$ index and introduce the notation ${}^{(j)}\mathbf{\mu}^M = E\big(\mathbf{p}^{M}_t\big)$ and ${}^{(j)}\mathbf{\mu}^B = E\big(\mathbf{p}^{M}_t\big)$. The vector ${}^{(j)}\mathbf{\mu}^M$ is the mean position for all points from the map that fall in voxel $j$.  The vector ${}^{(j)}\mathbf{\mu}^B$ is the mean position, in body-frame coordinates, for all current-scan points falling in voxel $j$. Substituting ${}^{(j)}\mathbf{\tilde{H}}_\mathcal{X}$, ${}^{(j)}\mathbf{\mu}^M$, and ${}^{(j)}\mathbf{\mu}^B$ into (\ref{eq:TaylorMu1}) gives the following equation.

\begin{equation}\label{eq:TaylorMu3}
    {}^{(j)}\mathbf{\mu}^M = {}^M\mathbf{\hat{R}}^{B}_t \;\; {}^{(j)}\!\mathbf{\mu}^B+ \mathbf{\hat{m}}^M_t+
    {}^{(j)} \mathbf{\tilde{H}}_\mathcal{X} \delta \mathcal{X}
\end{equation}

The state-correction vector $\delta \mathcal{X}$ is obtained by solving this equation simultaneously over all voxels. With a sufficient number of geometrically diverse voxels, the system is overdetermined, and a weighted least-squares solution can be used to mitigate measurement noise.

\begin{equation}\label{eq:wls}
    \delta \mathcal{X} = (\mathbf{H}^T \mathbf{W} \mathbf{H})^{-1} \mathbf{H}^T \mathbf{W} ~\mathbf{y}
\end{equation}

\noindent Here $\mathbf{W}$ is a weighting matrix, $\mathbf{H}$ is the concatenation of the matrices ${}^{(j)}\mathbf{\tilde{H}}_\mathcal{X}$ over all voxels $j$, and the  $\mathbf{y}$ is the concatenation of the observables ${}^{(j)}\mathbf{y}$ over the same voxels, where

\begin{equation}
{}^{(j)}\mathbf{y} = {}^{(j)}\mathbf{\mu}^M-{}^M\mathbf{\hat{R}}^{B}_t \;\; {}^{(j)}\!\mathbf{\mu}^B- \mathbf{\hat{m}}^M_t.
\end{equation}

\noindent The weighting matrix $\mathbf{W}$ is constructed from the measurement-error covariance for $\mathbf{y}$, as discussed in \parencite{2DICET}. The same reference also provides numerical tools that allow (\ref{eq:wls}) to be solved using a block decomposition of $\mathbf{H}$, $\mathbf{W}$, and $\mathbf{y}$. The decomposition enhances computational efficiency and reduces memory requirements.  

Once the correction states are computed, the initial guess $\hat{\mathcal{X}}$ can be updated using the following equation.

\begin{equation}\label{eq:updateChi}
    \hat{\mathcal{X}} \rightarrow  \hat{\mathcal{X}} + \delta \mathcal{X}
\end{equation}

\noindent Newton iteration can be used to converge on a solution.  In practice, we obtained better results using Levenberg-Marquardt optimiztion~\parencite{levenberg}.

%%%%%%%%%%%
%%%%%%%%%%%%% START COMMENTED SECTION %%%%%%%%%%%
%%%%%%%%%%%
\if false

In VICET, the relationship between $p^{B_0}$ and $p^M$ is represented by substituting parameters from the 6 state vector, $\mathbf{x}_0$ in to Eq. (\ref{eq:B02MapPluggedIn})

\begin{equation}\label{eq:B02MapPluggedIn}
    p^{M} = R(\phi_0, \theta_0, \psi_0) p^{B_0} + 
    \begin{bmatrix}
        x_0 \\
        y_0 \\
        z_0
    \end{bmatrix}
\end{equation}

Eq. (\ref{eq:Bend2B0}) allows the coordinates of a point in the frame of the sensor's location at the end of the sweep, $B_{end}$ to be represented in the frame of $B_0$.  

\begin{equation}\label{eq:Bend2B0}
    p^{B_0} = R^{B_{end}}_{B_0} p^{B_{end}} + t^{B_{end}}_{B_0}
\end{equation}

Visualised in Fig. (\ref{fig:fromulationFrames}), motion of the LIDAR sensor during a sweep causes each point in a scan to be recorded from a unique location of the sensor. We assume all points with a common $\alpha$ are recorded simultaneously, neglecting any internal bus delay between return channels in the sensor. 
% This change in frame during the scan is often neglected in LIDAR odometry tasks where the relative change in motion distortion between frames is minimal, however, as we demonstrate in our experiment this effect becomes severe in the map matching task. 
Applying a zero-order hold on the velocity of the sensor during scan $j$, we assume the location of the sensor while recording point $p$, $B_p$, can be linearly interpolated between $B_0$ and $B_{end}$ by scaling by the sweep angle of point $p$, by scaling factor $s_p$ which we define as $\frac{\alpha}{2\pi}$. Eq. (\ref{eq:Bp2B0PluggedIn}) relates a point's coordinates in $B_p$ to its coordinates in $B_0$.

\begin{equation}\label{eq:Bp2B0}
    p^{B_0} = R^{B_{p}}_{B_0} p^{B_{p}} + t^{B_{p}}_{B_0}
\end{equation}

\begin{equation}\label{eq:deltaXdef}
    \mathbf{\Delta x} = 
    \begin{bmatrix}
        \Delta x & \Delta y & \Delta z & \Delta \phi_0 & \Delta \theta_0 & \Delta \psi \\
    \end{bmatrix}
\end{equation}

\begin{equation}\label{eq:Bp2B0PluggedIn}
    p^{B_0} = R(s_p \Delta \phi, s_p \Delta \theta, s_p \Delta \psi) p^{B_p} + 
    \begin{bmatrix}
        s_p \Delta x \\
        s_p \Delta y \\
        s_p \Delta z
    \end{bmatrix}
\end{equation}

Combining Equations (\ref{eq:B02Map}) and (\ref{eq:Bp2B0}), a point $p$'s coordinates (which are captured in the frame of the sensor while recording $p$) can be represented in the Frame of the HD Map using the composite transform in Eq. (\ref{eq:Bp2M}). Combining $\mathbf{x_0}$ and $\mathbf{\Delta x}$ into a single 12-state solution vector $\mathcal{X}$, equation (\ref{eq:Bp2M}) can be simplified to operator $h$ which remaps a point from the body frame to the map frame. 

\begin{equation}\label{eq:Bp2M}
    p^M = R^{B_{end}}_{B_0}(R^{B_{p}}_{B_0} p^{B_{p}} + t^{B_{p}}_{B_0}) + t^{B_{0}}_{M} %=  h(j, \mathcal{X})
\end{equation}

\begin{equation}\label{eq:defh}
    p^M =  h(j, \mathcal{X})
\end{equation}

% \begin{equation}\label{eq:Bp2MPluggedIn}
%     \begin{tiny}
%     p^{M} = R(\phi_0, \theta_0, \psi_0) \bigg{(}R(s_p \Delta \phi, s_p \Delta \theta, s_p \Delta \psi) p^{B_p} + 
%     \begin{bmatrix}
%         s_p \Delta x \\
%         s_p \Delta y \\
%         s_p \Delta z
%     \end{bmatrix} \bigg{)}
%     + \begin{bmatrix}
%         x_0 \\
%         y_0 \\
%         z_0
%     \end{bmatrix}
%     \end{tiny}
% \end{equation}

VICET holds the reference scan $i$ constant and iteratively undistorts the raw cloud $j$ using an estimate of $\mathcal{X}$ (denoted as $\hat{\mathcal{X}}$) that minimizes the distance between corresponding features in each scan. 
Applying (\ref{eq:defh}) to all points in a raw point cloud, $j$, an estimate of the cloud in the frame of the HD Map can be represented by $j^M$

\begin{equation}\label{eq:j2jM}
    j^M = h(j, \hat{\mathcal{X}})
\end{equation}

After each iteration, correspondences are recalculated and $\hat{\mathcal{X}}$ is updated. This process repeats until the algorithm converges on a solution for $\hat{\mathcal{X}}$. Like NDT \parencite{D2DNDT}, VICET makes use of a voxel grid, and fits Gaussian distributions to the clusters of points from $i$ and $j^M$ inside each voxel to act as corresponding features. Let $^{(n)} \mu_{j^M} \in \mathcal{R}^3$ be the mean of points from $j^M$ inside voxel $n$, and $^{(n)} \Sigma_{j^M} \in \mathcal{R}^{3 \times 3}$ be their covariance.
Similarly, let $^{(n)} \mu_{i} \in \mathcal{R}^3$ be the mean of points from reference scan $i$ inside voxel $n$, and $^{(n)} \Sigma_{i} \in \mathcal{R}^{3 \times 3}$ be their covariance. Concatenating the difference between means from each voxel for scans $i$ and $j^M$ produces the residual vectors $\mathbf{y} \in \mathcal{R}^{N \times 3}$ shown in (\ref{eq:y}), where $N$ is the total number of occupied voxels.

\begin{equation}\label{eq:y}
    \mathbf{y} =  
    \begin{bmatrix}
        ^{(1)} \mu_{i} ~- ~^{(1)} \mu_{j^M} \\
        ^{(2)} \mu_{i} ~- ~^{(2)} \mu_{j^M} \\
        \vdots \\
        ^{(N)} \mu_{i} ~- ~^{(N)} \mu_{j^M}
    \end{bmatrix}
\end{equation}

VICET uses least-squares processing to estimate a linear perturbation $\Delta \mathcal{X}$ that minimizes the residual vector $\mathbf{y}$ according to equation (\ref{eq:lsdef}) where $H_{\mathcal{X}}$ is a matrix containing the appended partial derivatives of $\mathcal{X}$ with respect to point coordinates of each voxel.

\begin{equation}\label{eq:lsdef}
    \textbf{y} = \textbf{H}_{\mathcal{X}} \Delta \mathcal{X}
\end{equation}

When constructing $\mathbf{H}$, it is useful to consider how its structure consists of two parts: those relating to the $\mathbf{x}_0$ states (columns 1-6) and those relating to the $\Delta \mathbf{x}$ states (columns 7-12).

\begin{equation}\label{eq:defH}
    \mathbf{H}_{\mathcal{X}} = 
    \begin{bmatrix}
        % \mathbf{H}_{x_0} & \mathbf{H}_{\Delta x}
        ~^{(1)}\mathbf{H}_{\mathbf{x}_0} & ^{(1)}\mathbf{H}_{\Delta \mathbf{x}} \\
        ~^{(2)}\mathbf{H}_{\mathbf{x}_0} & ~^{(2)}\mathbf{H}_{\Delta \mathbf{x}} \\
        \vdots  & \vdots \\
        ~^{(N)}\mathbf{H}_{\mathbf{x}_0} & ~^{(N)}\mathbf{H}_{\Delta \mathbf{x}} \\
    \end{bmatrix}
    \in \mathcal{R}^{3N \times 12}
\end{equation}

$^{(n)}\mathbf{H}_{\mathbf{x}_0}$ describes how much small change in each parameter of $\mathbf{x}_0$ effects the distances in $[x,y,z]$ between distribution centers in voxel $(n)$.

\begin{equation}\label{eq:Hx0}
    {}^{(n)}\mathbf{H}_{\mathbf{x}_0} = 
    \Bigg{[}
    \begin{matrix}
        1 & 0 & 0 \\
        0 & 1 & 0 \\
        0 & 0 & 1 \\
    \end{matrix}
    \Bigg{|}
    \begin{matrix} 
        \frac{\delta R}{\delta \phi} ~{}^{(n)}\mu_{j^M}  & \frac{\delta R}{\delta \theta} ~{}^{(n)}\mu_{j^M} & \frac{\delta R}{\delta \psi} ~{}^{(n)}\mu_{j^M} 
    \end{matrix}
    \Bigg{]}
\end{equation}

Likewise, $^{(n)}\mathbf{H}_{\mathbf{\Delta x}}$ describes how much a small change in each parameter in $\mathbf{\Delta \mathcal{X}}$ effects the distances in $[x,y,z]$ between distribution centers in voxel $(n)$. Unlike (\ref{eq:Hx0}) where each block of $\mathbf{H}_{\mathbf{x_0}}$ contributes equally, each block of $\mathbf{H}_{\mathbf{\Delta x}}$ must be scaled by it's angular  placement in the sweep of the sensor. Recall, the $\mathbf{\Delta x}$ states describe the change in pose of the sensor between the beginning of the LIDAR sweep and recording a given point. If a considered voxel $(n)$ occupies a region in the beginning of the scan $(\alpha \approx 0)$, the difference in it's coordinates in $B_p$ and $B_0$ will be small, even if $\mathbf{\Delta x}$ is large. 

This scaling parameter is replaced with a simplified estimate in Eq. (\ref{eq:HdeltaX}), using the  mean sweep angle of the voxel mean from $j_M$, rather than mean sweep angle of all points...

\begin{equation}\label{eq:HdeltaX}
    {}^{(n)}\mathbf{H}_{\mathbf{\Delta x}} \approx \frac{{}^{(n)}\alpha_{j^M}}{2\pi}
    \Bigg{[}
    \begin{matrix}
        % 1 & 0 & 0 \\
        % 0 & 1 & 0 \\
        % 0 & 0 & 1 \\
        \hdots
    \end{matrix}
    \Bigg{|}
    \begin{matrix} 
        % \frac{\delta R}{\delta \phi} ~{}^{(n)}\mu_{j^M}  & \frac{\delta R}{\delta \theta} ~{}^{(n)}\mu_{j^M} & \frac{\delta R}{\delta \psi} ~{}^{(n)}\mu_{j^M} 
        \hdots
    \end{matrix}
    \Bigg{]}
\end{equation}

Similar to D2D-NDT \parencite{D2DNDT}, we weight each residual in $\mathbf{y}$ by the mahalanobis distance of the corresponding offset between distribution centers to arrive at Eq. (\ref{eq:wls}).

\begin{equation}\label{eq:wls}
    \Delta \mathcal{X} = (\mathbf{H}_{\mathcal{X}}^T \mathbf{W} \mathbf{H}_{\mathcal{X}})^{-1} \mathbf{H}_{\mathcal{X}}^T \mathbf{W} ~\mathbf{y}
\end{equation}

After solving for $\Delta \mathcal{X}$, the current approximation of $\hat{\mathcal{X}}$ is augmented. The new estimate of $\hat{\mathcal{X}}$ is then used to transform $j$ to $j_M$, distributions are refit to the new clusters of points within each voxel and the process repeats until convergence.

\begin{equation}\label{eq:updateChi}
    \hat{\mathcal{X}} \rightarrow  \hat{\mathcal{X}} + \Delta \mathcal{X}
\end{equation}

\fi
%%%%%%%%%%%
%%%%%%%%%%%%% END COMMENTED SECTION %%%%%%%%%%%
%%%%%%%%%%%

\subsection{Additional Details}
In this section we identify  practical details related to instantiating the above VICET algorithm in code.

%\begin{itemize}
    %\item 
\underline{Timestamps}: Timestamps of LIDAR points can be approximated by using beam angle $\psi$. It is important not to group points recorded at different times (e.g., at the beginning and end of the scan). We address the aliasing issue by defining a voxel boundary at $0^\circ$ (start of scan) and then removing occasional aberational points with $\psi < 0^\circ$ or $\psi \ge 360^\circ$. 
    % \item plays very nicely with spherical voxels in NDT
    %\item Because a gradient decent approach is used, VICET can converge into the wrong basin of attraction.  In our work, such instances are readily identified as wildly divergent from the initial guess. \hl{Bigger voxel more reliable?}.
    %\item 
    
\underline{Initialization}: As mentioned in connection with (\ref{eq:Taylor}), VICET assumes a reasonable estimate of initial pose. We obtained the initial pose by first running a standard scan-match, specifically NDT; VICET then improves this estimate by compensating for motion distortion.
    %\item 
    
\underline{Extended surfaces}: NDT fails to recognize that walls (and other flat surfaces stretching across voxel boundaries) provide useful information only in the surface-normal direction; to enhance convergence reliability and accuracy, we incorporate ICET \parencite{3DICET} extended-surface suppression into VICET.
%\end{itemize}

% \newpage

\section{Experiment: HD-Map Registration}

\begin{figure}[t]
\centering
\includegraphics[width=3.4in]{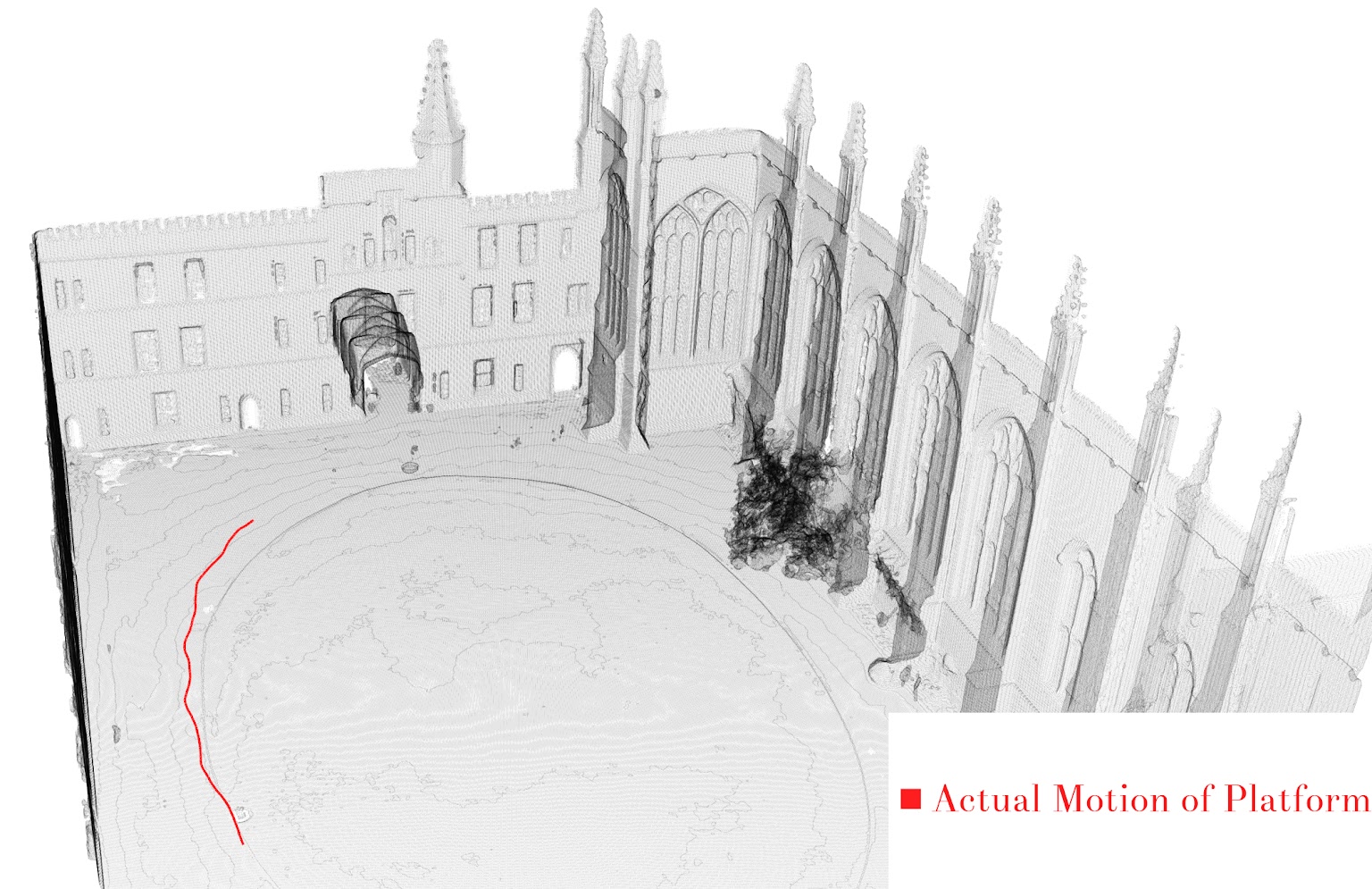} %Iuploaded one both red and black markers 
\caption{Ground truth for each of 200 scans, superimposed on the HD Map}
\label{fig:experiment1viz}
\end{figure}

% In this section we demonstrate how our proposed approach can achieve accurate estimate both the rigid transform and motion correction required to localize a LIDAR sensor against an  HD Map  using only a single raw point cloud.
In this section we evaluate the performance of snapshot scan-to-map localization using VICET, and we compare with two other widely used snapshot algorithms: NDT and ICP.
For our evaluation, we make use of the \textit{Newer College Dataset} \parencite{newercollege} which contains a rosbag of raw LIDAR data packets and associated ground truth.  The dataset also includes a survey-grade HD map of the structured environment recorded on a separate Leica total station.
% Can't use LOAM (which uses motion historty to correct distortion) since it requires a historical model of vehicle motion and will not work with a single scan.
Specifically, we utilize 200 consecutive LIDAR scans of Newer College Data from the \textit{Quad With Dynamics} trajectory, where a researcher runs while swinging the LIDAR sensor at head height, to approximate the dynamics of a multi-rotor drone \parencite{newercollege}. The ground-truth position for the 200 scans is shown in Fig. (\ref{fig:experiment1viz}), superposed on the HD map. 

Snapshot processing was conducted by matching each individual scan to the map, without reference to any other scan from the sequence.
Snapshot position and orientation estimates were obtained via three distinct methods: VICET, ICP, and NDT. Note that VICET was initialized with NDT, and so the same spherical voxel grid \parencite{SphericalICET} was used for both VICET and NDT in order to ensure a controlled comparison. The three methods were compared using absolute-error and chamfer-distance
% ~\parencite{van2000image, lu2022oriented} 
performance metrics. 

\begin{figure}[t]
\centering
% old graphics, all on one plot
% \includegraphics[width=3.5in]{graphics/scanToHDMapError.jpg}
% \includegraphics[width=3.5in]{graphics/scanToHDMapYawError.jpg}
%new graphics, plotted side by side
% \includegraphics[width=3.4in]{graphics/scanToHDMapErrorV2.jpg}
% \includegraphics[width=3.35in]{graphics/scanToHDMapYawErrorV2.jpg}
\includegraphics[width=3.4in]{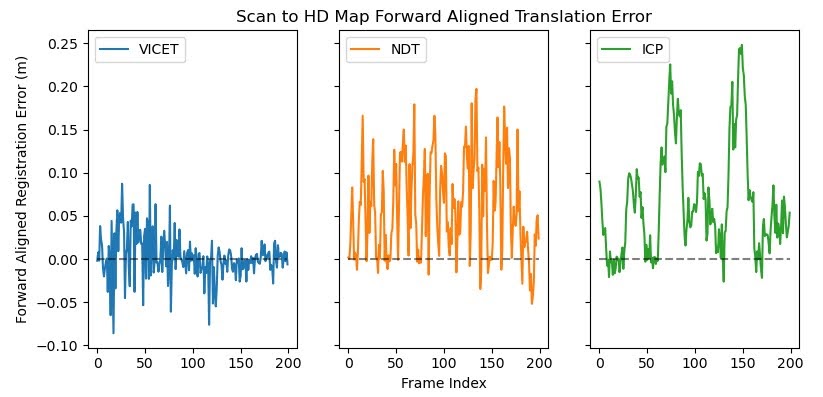}
\includegraphics[width=3.35in]{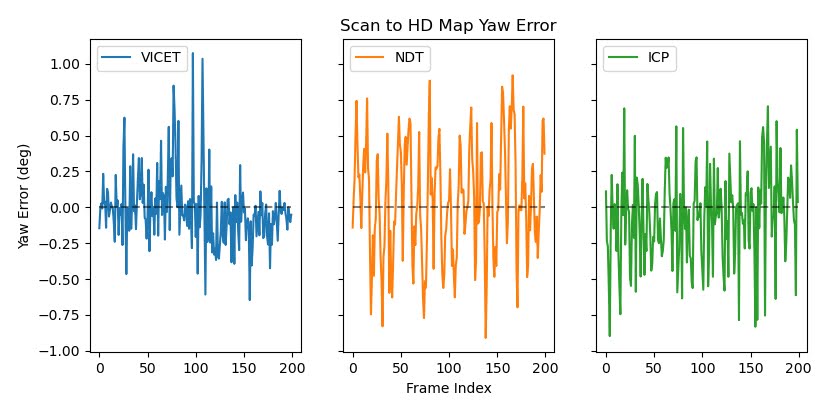}
\caption{Translation error (top) and yaw error (bottom) when registering 200 raw LIDAR scans against a static HD Map. }
\label{fig:Scan2HDMap}
\end{figure}

\subsection{Absolute-Error Results}

An absolute-error vector was formed from each estimate by subtracting ground-truth values, which were referenced to LIDAR position and orientation at the start of the scan.
% This sequence was selected to minimize attraction between raw scan data and non-line-of-sight points in the HD Map.
We display error values in Fig. (\ref{fig:Scan2HDMap}). Specifically the figure shows forward-translation error (top) and yaw-angular error (bottom) for each of the three algorithms under comparison. Forward error was typically the largest component of the translation-error vector.  By contrast with translation errors, angular errors did not exhibit directional dependence, so we selected as representative the angular-error component in the yaw sense (about the map-fixed vertical axis).

Absent motion-distortion correction, the forward-errors for NDT and ICP exhibit a significant bias.  As evident in Fig. (\ref{fig:Scan2HDMap}), errors for both NDT (orange) and ICP (green) are generally above zero, with very few negative values.  By comparison, the corresponding VICET forward-error values (blue) appear to be zero-centered. Additionally, the distribution of the VICET forward-error values is narrower than the distributions for NDT (orange) or ICP (green). These trends were quantified by computing the mean and standard deviation of the forward-error values across each of the 200 samples, statistics that are reported in Table (\ref{tab:meanAndSTD}). The table confirms the statistical mean of forward error for VICET is an order of magnitude smaller than for uncompensated ICP or NDT (0.269 cm as compared to more than 6.0 cm); similarly, the standard deviation of forward error for VICET is better than half that for ICP or NDT (2.6 cm as compared to more than 5.4 cm). 

Similar trends are observed in yaw, with motion-corrected VICET outperforming the uncorrected NDT and ICP algorithms.  Again, VICET exhibits a substantially smaller bias ($0.015^\circ$ compared to $0.05^\circ$ or more) and a slightly narrower standard deviation ($0.24^\circ$ as compared to $0.31^\circ$ or more).

%Error in the forward component of NDT (orange) and ICP (green) solutions jump between values near zero and peaks between 15cm and 20cm. 
% This is because ICP and NDT can only solve for a rigid transform to align the distorted cloud with the undistorted keyframe cloud. 
%This is because in NDT and ICP, the forward motion of the platform during each scan (between 12 cm and 15 cm per frame) stretches the LIDAR scans along the direction of motion-- localization error is introduced when rigid algorithms attempt to fit the stretched side of the point cloud at the end of a scan (where $t \approx T$) against the corresponding undistorted %reference feature in the HD Map.
%By solving for distortion states, VICET is able to compensate for this relative stretching between corresponding features and produce estimates with zero-centered error. VICET is also able to achieve a similar, albeit less stark, improvement in yaw as it is able to "unpinch" the raw scans to match the reference cloud as shown in Fig. (\ref{fig:beforeVsAfter}). 

\if false
%Old table- RMSE
\begin{table}
    \caption{Localization Error} 
    \centering
    \setlength{\tabcolsep}{0.3\tabcolsep}
    \begin{tabularx}{0.48\textwidth} {  %was 0.48\textwidth
      | >{\centering\arraybackslash}X
      | >{\centering\arraybackslash}X
      | >{\centering\arraybackslash}X
      | >{\centering\arraybackslash}X | }
    \hline
     & RMS Forward Translation Error (cm)  & RMS Yaw Error (deg) \\
    \hline
    ICP & 9.269 & 0.317\\
    \hline
    NDT & 8.097 & 0.389 \\ 
    \hline
    \textbf{VICET (ours)} & \textbf{2.626} & \textbf{0.242} \\ 
    \hline
    \end{tabularx}
    \label{tab:RMS}
\end{table} 
\fi

%alt table - Mean and STD 
\begin{table}
    \caption{Localization Error} 
    \centering
    \setlength{\tabcolsep}{0.3\tabcolsep}
    \begin{tabularx}{0.48\textwidth} {  %was 0.48\textwidth
      | >{\centering\arraybackslash}X
      | >{\centering\arraybackslash}X
      | >{\centering\arraybackslash}X
      | >{\centering\arraybackslash}X
      | >{\centering\arraybackslash}X
      | >{\centering\arraybackslash}X | }
    \hline
    \multirow{2}{*}{ } & \multicolumn{2}{c|}{Forward Translation Error}  & \multicolumn{2}{c|}{Yaw Error} \\
    \cline{2-5} & Mean (cm) & STD (cm) & Mean (deg) & STD (deg) \\
    \hline
    ICP &  6.671  & 6.435  & -0.0657 & 0.310\\
    \hline
    NDT & 6.005 & 5.432 & 0.0509 & 0.386 \\ 
    \hline
    \textbf{VICET} & \textbf{0.269} & \textbf{2.612} & \textbf{-0.015} & \textbf{0.242} \\ 
    \hline
    \end{tabularx}
    \label{tab:meanAndSTD}
\end{table} 

\begin{table}
    \caption{Normalized Chamfer Distance ($\text{cm}^2$)} 
    \centering
    \setlength{\tabcolsep}{0.3\tabcolsep}
    \begin{tabularx}{0.48\textwidth} {  %was 0.48\textwidth
      | >{\centering\arraybackslash}X
      | >{\centering\arraybackslash}X
      | >{\centering\arraybackslash}X
      | >{\centering\arraybackslash}X | }
    \hline
     & \textit{Quad With Dynamics} & \textit{Dynamic Spinning} \\
    \hline
    ICP & 0.869 & 3.317\\
    \hline
    NDT & 0.814 & 3.045\\ 
    \hline
    \textbf{VICET} & \textbf{0.732} & \textbf{2.118} \\
    \hline
    \end{tabularx}
    \label{tab:chamfer}
\end{table}

\subsection{Chamfer Distance}

\begin{figure}
\centering
\includegraphics[width=3.0in]{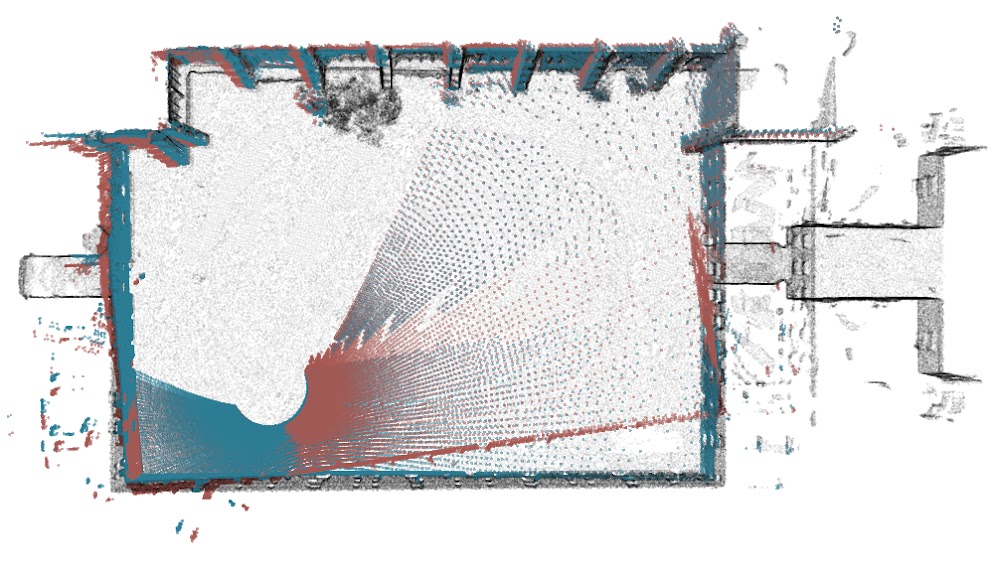} %top down view
\caption{Registered point clouds with and without motion-distortion compensation. As compared to the NDT-registered data (red), the VICET corrected data (blue) more closely resemble the HD-Map (black).}
\label{fig:beforeVsAfter}
\end{figure}

%We perform an additional experiment where we quantify registration error in terms of chamfer distance, which we report in Table (\ref{tab:chamfer}).
We used a second metric, chamfer distance, to directly compare the shape of the map to that of the current scan (after registration). Chamfer distance is defined as the sum of the squared distances for all nearest-point correspondences between two clouds.  This metric effectively compares the different shapes of registered scans visualized in Fig. (\ref{fig:beforeVsAfter}), as referenced to the map. A larger shape difference corresponds to a larger chamfer distance. Notably, the chamfer distance metric does not rely on ground truth.

Some minimal data cleaning was required to compare chamfer distance in a meaningful way. As a first step, ICP, NDT and VICET were used to register a scan to the map. A chamfer distance was then computed applying an outlier rejection criterion (no points considered outside the inflated convex hull for the map, with 5\% applied in our analysis).  The process was repeated for all 200 scans considered from the \textit{Quad with Dynamics} trajectory.  Each chamfer distance was then normalized by the number of points in the associated LIDAR cloud, to account for scan-to-scan variability in the number of non-return points.  Normalized chamfer distances were then averaged across all 200 cases. The results are compiled in Table (\ref{tab:chamfer}).

The table (column labeled \textit{Quad with Dynamics}) shows that the average normalized chamfer distance for VICET was slightly better than for either NDT or ICP (0.73 cm$^2$ as compared to 0.81 cm$^2$ or more).  This result indicates that the output from VICET is indeed a better approximations of the map than the outputs of either NDT or ICP.  

Because these chamfer distances were only modestly separated, we evaluated a second Newer College trajectory, labeled \textit{Dynamic Spining}, to verify the dependence of chamfer distance on distortion severity. The second trajectory contained more motion distortion than the first. Normalized chamfer-distance results were computed and the data compiled, as recorded in the table (column labeled \textit{Dynamic Spinning}). As expected, the chamfer distances were higher for this case than for the prior case (\textit{Quad with Dynamics}).  Moreover, the chamfer distance
metric for VICET was again better than for either NDT or ICP (2.1 cm$^2$ as compared to 3.0 cm$^2$ or more).

\section{Discussion and Future Work}

The experimental results from the prior section strongly support the importance of motion-distortion correction for high-accuracy scan-to-map matching. 
%We showed VICET accomplishes distortion correction using only a single scan and a map. Absent correction, motion distortion introduces systematic biases in estimated pose, as evidenced by Table \ref{tab:meanAndSTD}.
Distortion not only introduces systematic biases but also inflates the standard deviation, implying that uncertainty quantification methods (e.g., estimators predicting state-error covariance) will perform poorly without distortion-mitigation. The higher standard deviation (and wider error distribution) for uncompensated algorithms may possibly be explained by the registration-process ``hunting'' over the space of poses observed between the start and end of the scan ($0\le t \le T$); those intermediate poses are all partially represented in the raw scan.

In this work, our primary emphasis is on achieving high-accuracy scan-to-map matching with a reasonable initial guess.  
VICET, by itself, does not solve the \textit{Kidnapped Robot} problem \parencite{thrun2002probabilistic},
however, VICET could be layered with other algorithms that solve the global search problem, enhancing their accuracy.  In the simplest form, VICET might be run to convergence from a number of initial guesses 
%(for $\mathbf{x_0}$ and $\mathbf{\Theta_0}$) 
using an exhaustive search, a particle filter,  
%of obtaining initial registration might be 
or through a pose-invariant ``global'' registration network such as \textit{DeepGMR} \parencite{deepgmr} 
%to align the distorted clouds 
before compensating for distortion with VICET. 

Though not developed for LIDAR odometry (LO), VICET might still have possible future LO applications. 
% The potential benefit is that 
VICET provides immediate unwarping when comparing a new scan to an LO submap, without delay. By comparison, many existing LO algorithms infer unwarping in one scan by estimating the distance traveled from that scan to the next, a process which inherently adds a delay (typically 50-200 ms to capture the next scan). Such delays are undesirable for feedback-control loops, as pure delays cause sluggish performance and erode stability margins.

%. The algorithm introduced in this paper provides a way to accurately estimate position of a platform using only an HD-Map and a single distorted LIDAR scan. Using a single scan to solve for both position and distortion states allows the VICET algorithm to begin processing immediately after a sweep of the sensor is completed. Existing techniques that account for motion distortion either rely on external data \parencite{in2lama, videocorrection} which may not be available, or must wait for the sensor to record a second scan \parencite{LOAM, vicp} which introduces unnecessary delay in the localization process.  

%Like other "local" scan registration algorithms, reliable convergence of VICET is dependant on an initial estimate of the solution vector falling within the correct basin of attraction. 
% In our experiments on the \textit{Newer College Dataset}, the walls of the courtyard are largely visible for all scans, so the basin of attraction is sufficiently large to ensure convergence, so long as the initial rotation of the sensor $\ll 90^{\circ}$ from the correct value along each axis. 

%In our experiments, the frames selected along the South end of the courtyard align within $45^{\circ}$ of each body-frame axis of the LIDAR sensor, so a sweep through multiple initial angles similar to \parencite{magnusson2009three} was not required.

To apply VICET in larger, more complex, and occluded environments, it may be necessary to address sensitivity to perspective shift bias \parencite{perspectiveShift}. 
Perspective bias can occur in HD-Map matching when LIDAR points exist in the map but are not visible from the location of the current scan, because they are occluded by foreground surfaces. 
% These points in the HD Map are visible from other perspectives, just not the current one.
New methods are needed to condition voxel-based processing to minimize systematic errors due to perspective effects.  Fortunately, in the \textit{Newer College Dataset}, the walls of the courtyard are mostly visible from all vantage points, 
%, so the basin of attraction is largely constrained by the initial orientation of the sensor relative to the HD Map. 
%The algorithm does not account for perspective-shift bias, a systematic error related to viewpoint \parencite{perspectiveShift}, occurs because 
so perspective errors did not play a major role in this analysis.

\section{Conclusion}

In this paper we introduced VICET, a novel algorithm to solve the scan-to-map matching problem for a single LIDAR point cloud.  VICET soves for twelve states, six describing a rigid transform aligning the scan to the map, and six more to account for distortion due to platform motion during the creation of the cloud. In contrast with other motion-distortion methods, our approach requires only a single LIDAR scan and no external sensor data.

Through experiments on real-world data, we demonstrate VICET improves accuracy over conventional NDT and ICP, 
%how our technique provides a significant accuracy improvement over widely used scan-to-map registration techniques that do not perform distortion correction. Using a trajectory drawn from the Newer College dataset, we showed that VICET reduced 
reducing the translation-bias for map matching (by an order of magnitude, from 6.9 cm to 0.27 cm) while reducing one-sigma variations (from 5.4 cm to 2.6 cm). VICET also reduced bias and variance for orientation estimation. These enhancements are relevant for   precision automotive and urban air mobility applications. 

%\hl{Correcting all sources of error in scan registration can permit a safe application of autonomous systems to highly dynamic applications such as drone flight and autonomous vehicles.} 

%Our proposed method accomplishes distortion correction using only a single scan and a map, in contrast with existing LIDAR odometry algorithms, like LOAM and VICP, which rely on a sequence of many scans to identify and mitigate motion distortion

\section*{Acknowledgments}
The authors gratefully acknowledge the Tufts graduate funding that made this work possible. The authors also acknowledge and thank the U.S. Department of Transportation Joint Program Office (ITS JPO) and the Office of the Assistant Secretary for Research and Technology (OST-R) for partial sponsorship. Opinions discussed here are those of the authors and do not necessarily represent those of the DOT or affiliated agencies.

\printbibliography

\if false
\appendix

\subsection{Structure of $\mathbf{H}$}

Let ${}^{(j)}\!\mathbf{\mu}^{B \rightarrow M}$ be the mean of the subset of points from the new scan that fall inside voxel $j$ after being brought to the Map frame with $\hat{\mathcal{X}}$

\begin{equation}
   {}^{(j)}\!\mathbf{\mu}^{B \rightarrow M} = {}^M\mathbf{\hat{R}}^{B}_t \;\; {}^{(j)}\!\mathbf{\mu}^B+ \mathbf{\hat{m}}^M_t
\end{equation}

The matrix $\mathbf{H}$, used for least-squares processing in Eq. (\ref{eq:wls}), is the vertical concatenation of submatrices ${}^{(j)}\mathbf{\tilde{H}}_\mathcal{X}$, each corresponding to the Jacobian of the estimated state vector $\hat{\mathcal{X}}$ with respect to ${}^{(j)}\!\mathbf{\mu}^{B \rightarrow M}$.

% When constructing $\mathbf{H}$, it is useful to consider how its structure consists of two parts: those relating to the $\mathbf{x}_0$ states (columns 1-6) and those relating to the $\Delta \mathbf{x}$ states (columns 7-12).

\begin{equation}\label{eq:defH}
    \mathbf{H} = 
    \begin{bmatrix}
        % ^{(1)}\tilde{\mathbf{H}}_{\mathcal{X}} & ^{(2)}\tilde{\mathbf{H}}_{\mathcal{X}} & \hdots & ^{(J)}\tilde{\mathbf{H}}_{\mathcal{X}} \\
        ^{(1)}\tilde{\mathbf{H}}_{\mathcal{X}} \\
        ^{(2)}\tilde{\mathbf{H}}_{\mathcal{X}} \\
        \vdots \\
        ^{(J)}\tilde{\mathbf{H}}_{\mathcal{X}} \\
    \end{bmatrix}
    \in \mathcal{R}^{3J \times 12}
\end{equation}

Each submatix ${}^{(j)}\mathbf{\tilde{H}}_\mathcal{X}$ can be further decomposed into its constituent parts, which describe how much a small change in the respective parameter of $\hat{\mathcal{X}}$ effects the $[x,y,z]$ translation of ${}^{(j)}\!\mathbf{\mu}^{B \rightarrow M}$.
Recall, the goal of the optimization routine inside VICET is to minimize the offsets between the centers of distributions from the map and new LIDAR scan within each voxel.

\begin{equation}\label{eq:defHj}
     {}^{(j)}\mathbf{\tilde{H}}_\mathcal{X} = 
     \begin{bmatrix}
          {}^{(j)}\mathbf{\tilde{H}}_{\mathbf{x_0}} & {}^{(j)}\mathbf{\tilde{H}}_{\mathbf{\Delta x}} & {}^{(j)}\mathbf{\tilde{H}}_{\mathbf{\Theta_0}} & {}^{(j)}\mathbf{\tilde{H}}_{\mathbf{\Delta \Theta}} \\
     \end{bmatrix}
     \in \mathcal{R}^{3 \times 12}
\end{equation}

With respect to the rigid translation parameters $\mathbf{x_0}$, the corresponding section of the Jacobian is simply an identity matrix because a translation of the new scan directly maps to a rigid translation of distribution centers in voxel $j$. 

\begin{equation}\label{eq:defX0}
    {}^{(j)}\mathbf{\tilde{H}}_{\mathbf{x_0}}  = 
    \begin{bmatrix}
        1 & 0 & 0 \\
        0 & 1 & 0 \\
        0 & 0 & 1 \\
    \end{bmatrix}
\end{equation}

For translation distortion parameters $\mathbf{\Delta x}$, the process is similar however it requires values to be scaled by the time of ${}^{(j)}\!\mathbf{\mu}^{B \rightarrow M}$, as distortion grows linearly throughout the scan period. Here, ${}^{(j)}s$ represents the normalized time between the beginning and end of the scan for voxel $j$, as described in Section III. 

\begin{equation}\label{eq:defHDeltaX}
    {}^{(j)}\mathbf{\tilde{H}}_{\mathbf{\Delta x}}  = 
    \begin{bmatrix}
        ^{(j)}s & 0 & 0 \\
        0 & ^{(j)}s & 0 \\
        0 & 0 & ^{(j)}s \\
    \end{bmatrix}
\end{equation}

The component of ${}^{(j)}\mathbf{\tilde{H}}_\mathcal{X}$ from rigid rotation, $\mathbf{\Theta_0}$ can be calculated by dotting the partial derivatives of $f$ (as described in Eq. \ref{eq:RotMatrix}) with respect to each of the three Euler angels in $\mathbf{\Theta_0}$ with ${}^{(j)}\!\mathbf{\mu}^{B \rightarrow M}$

\begin{equation}\label{eq:defHTheta0}
    {}^{(j)}\mathbf{H}_{\mathbf{\Theta_0}} = 
    \begin{bmatrix} 
        \frac{\delta f}{\delta \mathbf{\Theta_{0,1}}} ~{}^{(j)}\!\mathbf{\mu}^{B \rightarrow M}  & \frac{\delta f}{\delta \mathbf{\Theta_{0,2}}} ~{}^{(j)}\!\mathbf{\mu}^{B \rightarrow M} & \frac{\delta f}{\delta \mathbf{\Theta_{0,3}}} {}^{(j)}\!\mathbf{\mu}^{B \rightarrow M} 
    \end{bmatrix}
    \in \mathcal{R}^{3 \times 3}
\end{equation}

The final component of ${}^{(j)}\mathbf{\tilde{H}}_\mathcal{X}$ associated with rotatinal distortion takes a similar form to Eq. (\ref{eq:defHTheta0}), however, like Eq. (\ref{eq:defHDeltaX}) ${}^{(j)}\mathbf{H}_{\mathbf{\Delta \Theta}}$ must also linearly scale the voxel's contribution by normalized time ${}^{(j)}s$ 

\begin{equation}\label{eq:defHDeltaTheta}
    {}^{(j)}\mathbf{H}_{\mathbf{\Delta \Theta}} = 
    \begin{bmatrix} 
        {}^{(j)}s\frac{\delta f}{\delta \mathbf{\Delta \Theta_{1}}} ~{}^{(j)}\!\mathbf{\mu}^{B \rightarrow M}  & {}^{(j)}s\frac{\delta f}{\delta \mathbf{\Delta \Theta_{2}}} ~{}^{(j)}\!\mathbf{\mu}^{B \rightarrow M} & {}^{(j)}s\frac{\delta f}{\delta \mathbf{\Delta \Theta_{3}}} {}^{(j)}\!\mathbf{\mu}^{B \rightarrow M} 
    \end{bmatrix}
    \in \mathcal{R}^{3 \times 3}
\end{equation}
\fi 

\end{document}